\documentclass[twocolumn]{jsiamletters}
\usepackage{graphicx}

\group{Applied Chaos
}  

\affiliation{Graduate School of Engineering, Fukuoka Institute of Technology}{Wajiro 3-30-1, Higashi-ku, Fukuoka, 811-0295 Japan}
\affiliation{Faculty of Information Engineering, Fukuoka Institute of Technology}{Wajiro 3-30-1, Higashi-ku, Fukuoka, 811-0295 Japan}
\authorinfo{Yuki Tanaka}{1}{}
\authorinfo{Yutaka Yamaguti}{2*}{y-yamaguchi@fit.ac.jp}
\email{y-yamaguchi@fit.ac.jp}

\title{Evaluating generation of chaotic time series by convolutional generative adversarial networks}

\abstract{To understand the ability and limitations of convolutional neural networks to generate time series that mimic complex temporal signals, we trained a generative adversarial network consisting of convolutional networks to generate chaotic time series and used nonlinear time series analysis to evaluate the generated time series. A numerical measure of determinism and the Lyapunov exponent showed that the generated time series well reproduce the chaotic properties of the original time series. However, error distribution analyses showed that large errors appeared at a low but non-negligible rate. Such errors would not be expected if the distribution were assumed to be exponential.
}%
\keywords{chaos, generative adversarial network, convolutional network, nonlinear time series analysis}%

\begin{document}

\maketitle

\section{Introduction}

The generation and prediction of nonlinear time-series data are being actively studied as an application of dynamical systems theory~\cite{frank2001,kantz2003,jaeger2004}. Conventional approaches such as using feed-forward and recurrent neural networks (RNNs)~\cite{frank2001,jaeger2004} aim to find a deterministic transition rule 
\begin{equation}
  x_{n+1}=f(x_{n}) \label{eqn-f}
\end{equation}
inherent in the time series data and to iteratively generate time series using an approximation of $f$. 
Because of their deterministic but unstable nature, chaotic signals are often used to evaluate model performance on time-series generation and prediction tasks in machine learning~\cite{gilpin2021}.
 
Convolutional neural networks (CNNs) that generate time-series samples without iterative computation in the time direction have recently been developed and applied to many time-series generation tasks, such as speech synthesis~\cite{donahue2018,zhang2022}. 
Because such CNNs can generate time series non-causally (i.e., without auto-regressive iterations of Eq.~\eqref{eqn-f}), they can leverage the power of GPU-based parallel computing to rapidly generate many time series.
CNNs outperform RNN-based models for many sequential tasks in terms of speed and computational cost~\cite{bai2018}. Time-series data generated by generative adversarial networks (GANs)~\cite{goodfellow2014} using convolutional networks~\cite{radford2015} are generally of high quality~\cite{donahue2018}.

An interesting question is whether these CNN-based models, which do not involve temporal iteration, can generate chaotic time series, and if so, to what extent the generated time series are chaotic. To determine which characteristics are preserved and which are lost in time series generated by deep generative models, we compared the properties of GAN-generated time series with those of the original time series. In particular, we examined whether two properties inherent to chaotic nonlinear time series, namely, their deterministic properties and sensitivity to initial values, were reproduced in the generated data. We also observed distributions of transition errors and investigated their specific properties.

%ADD for revision
Our objective does not aim to present the optimal model for accurately generating chaotic time series. Instead, our focus lies in exploring the nature of inaccurate approximations and inherent failures that arise during the learning of time series.
 Our contributions to the field of nonlinear dynamical systems and machine learning in this paper can be summarized as follows:
1) we conducted a comprehensive analysis of the accuracy and limitations of approximations from the viewpoint of nonlinear time series analysis, and identified anomalous error distributions.
2) We were able to provide information on the capabilities and limitations of GANs to researchers who use nonlinear time series generated by GANs for some downstream task applications.  
3) We discovered properties that could potentially be used to discriminate true time series from generated time series.
%% end mod

\section{Models}

GANs comprise two networks: a generator~$G$ and a discriminator~$D$~\cite{goodfellow2014}. Adversarial training between $G$ and $D$ enables $G$ to generate high-quality data similar to data in the training set. The generator learns a mapping from the latent space $Z$ to the data space $X$ in which the training samples are distributed. The discriminator evaluates whether the input data can be considered a sample from the training set by estimating the probability that the input came from the training data rather than generated by $G$. One-dimensional convolutional layers have often been used for $G$ and $D$ in sequence generation tasks~\cite{donahue2018}.

The objective function $V(G, D)$ used for GAN training is 
\begin{multline}
V(G,D) = \mathbb{E}_{x_{\mathrm{data}} \sim p_{\mathrm{data}}(x)}\left[\log D(x_{\mathrm{data}})\right] \\
+ \mathbb{E}_{z \sim p_{z}(z)}\left[\log\left(1-D(G(z))\right)\right],
\end{multline}
where $x_{\mathrm{data}} \sim p_{\mathrm{data}}(x)$ is a data sample drawn from a distribution of training data $p_{\mathrm{data}}$, 
$z \sim p_{z}(z) $ is a variable sampled from the latent space according to its probability distribution $p_{z}(z)$,
$D(x_{\mathrm{data}})$ is the output of the discriminator for input as a training sample,
$G(z)$ is the generated data output by $G$, and 
$D(G(z))$ is the discriminator output for the generated data.
The GAN training is formulated as a minimax game: $\min_{G}\max_{D}V(G, D)$.
In the implementation, the following two steps are performed iteratively: (i)~the discriminator learns to maximize $V(G, D)$, and (ii)~the generator learns to minimize $V(G, D)$.

In this experiment, we constructed $G$ and $D$ using a one-dimensional convolutional network. The model implementation is available from our GitHub repository~\cite{github}.
Briefly, $G$ was a model with nine weighted convolution layers, while $D$ was a model with seven weighted convolution layers and one dense layer as the final layer after a global average pooling layer. Both models employed a fully convolutional architecture\cite{radford2015} so that the length of the output time series of $G$ could be changed by changing the input size.
Instead of a transposed convolution layer, we used simple up-sampling layers to double the sequence size between the convolutional layers in $G$. We used the exponential linear unit function~\cite{clevert2015} as the activation function, with no regularization. 

The training objective was to reproduce the time series of the logistic map $f(x) = a x(1-x)$ with $a=4.0$. In each training session, trajectories of length $1098$ were computed by iteratively applying $x_{n+1} = f(x_n)$ from $100$ randomly determined initial values. We repeated 200,000 training cycles with such sets of $100 \times 1098$ training data per training session. 
All models were trained on NVIDIA GPUs with 32-bit floating-point precision.

\section{Results}
\subsection{Training results}

\begin{figure}[t]
  \centering
  \includegraphics[scale=0.25]{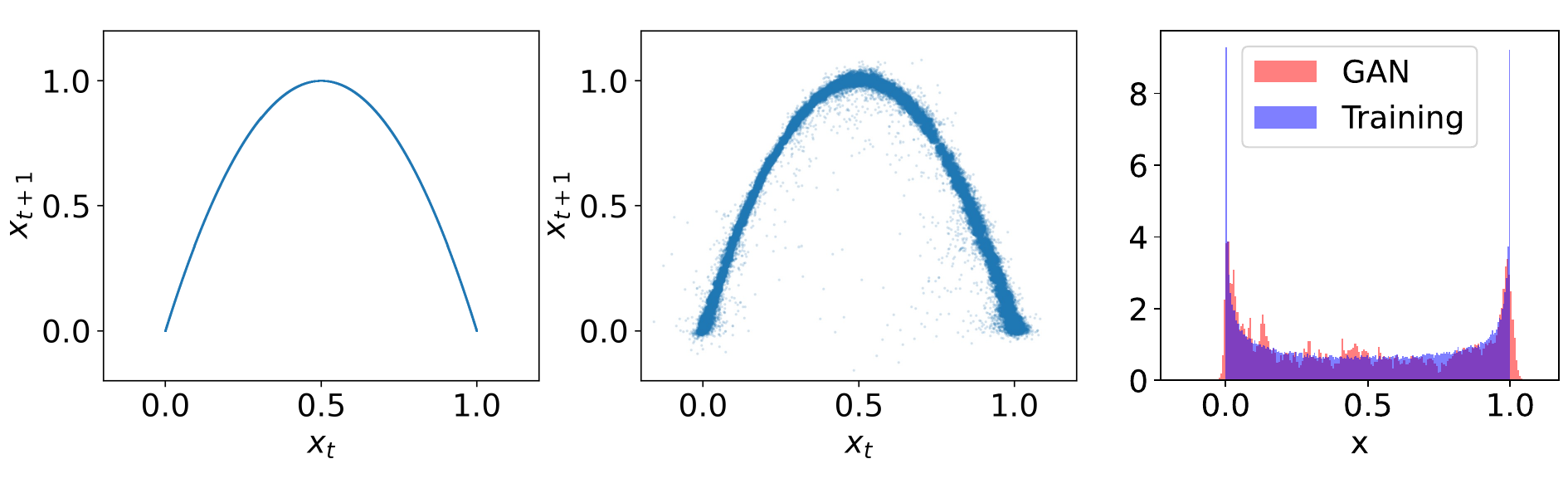}
  \caption{ Return maps of (left)~training data and (center)~generated data, and (right)~a histogram of $x_{t}$.}
  \label{fig-returnmap}
\end{figure}

Figure~\ref{fig-returnmap} shows return maps of training time series data (left) and generated time series data (center).
Both show 100,000 points.
As the figure shows, the return map for generated data well approximates that for training data. 
However, there are some large errors, statistics for which we present in Sec.~\ref{sec-errors}. On the right side of the figure is a histogram of $x_{t}$ values. Although the distribution of the generated data captures the training data outline, to some extent the distribution falls beyond the $[0,1]$ range.

\subsection{Deterministic properties}

We tested whether the generated data have deterministic properties, that is, 
to what extent a deterministic process such as Eq.~\eqref{eqn-f} explains the temporal transitions of variables. We used Wayland's method~\cite{wayland1993}, which measures determinism as the phase space continuity observed in the time series. 
We applied a surrogate method~\cite{theiler1992} to statistically evaluate the results of Wayland's method. 
Using the generated time series and the training time series, we constructed delay-embedding vectors with embedding dimension $m(=1,\ldots,5)$ as
\begin{equation*}
  \mathbf{x}_t = (x_t, x_{t+1},\ldots, x_{t+m-1}) \in \mathbb{R}^m, (t=1,..., L-m+1), 
\end{equation*}
where $L$ is the time-series length.
Then for vector $\mathbf{x}_{t_{0}}$ at time $t_{0}$, we identified $K$ ($=50$ in this experiment) nearest-neighbor points $\mathbf{x}_{t_1},\ldots, \mathbf{x}_{t_K}$ and computed the translation vector $\mathbf{v}_{t_i}=\mathbf{x}_{t_{i}+1}-\mathbf{x}_{t_i}$ for each $t_{i}$. We then calculated the translation error $E_{\text{trans}}$ as
\[
  E_{\text{trans}} = \frac{1}{K+1}\sum_{i=0}^{K} \frac{\left\Vert \mathbf{v}_{t_{i}} - \hat{\mathbf{v}} \right\Vert}{\left\Vert \hat{\mathbf{v}} \right\Vert},
  \]
  where $\hat{\mathbf{v}}= \frac{1}{K+1}\sum_{i=0}^{K}{\mathbf{v}_{t_i}}$. 
  We performed this calculation 1000 times using different reference times $t_{0}$, and evaluated determinism of the time series by the median $\bar{E}_{\mathrm{trans}}$ of $E_{\mathrm{trans}}$.  
For time series with mixed deterministic and stochastic properties, such as low-dimensional chaos with noise, $E_{\mathrm{trans}}$ becomes closer to $0$ as the noise decreases and determinism increases.

We used the surrogate data method~\cite{theiler1992} to statistically evaluate $\bar{E}_{\text{trans}}$.
This method generates many random time series that retain certain properties of the original time series, then tests whether the statistical properties of the original time series differ from those of the random time series.
We generated surrogate time series from the training time-series data using three algorithms: random shuffle, which randomly changes variable orders, Fourier transform, which randomly changes the phase without changing the amplitude of the original time series' Fourier transform, and amplitude-adjusted Fourier transform, which shuffles the phase while preserving the value distribution of the original data.
To compare GAN-generated time series with deterministic time series with added observational noise, we generated a ``noisy'' time series by adding Gaussian white noise to the training time series.
We set the variance of the observational noise to the squared mean of the transition error $e_{t}= f(x_{t})-x_{t+1}$ of the GAN-generated time series (about $0.0014$).

\begin{figure}[t]
  \centering
  \includegraphics[scale=0.4]{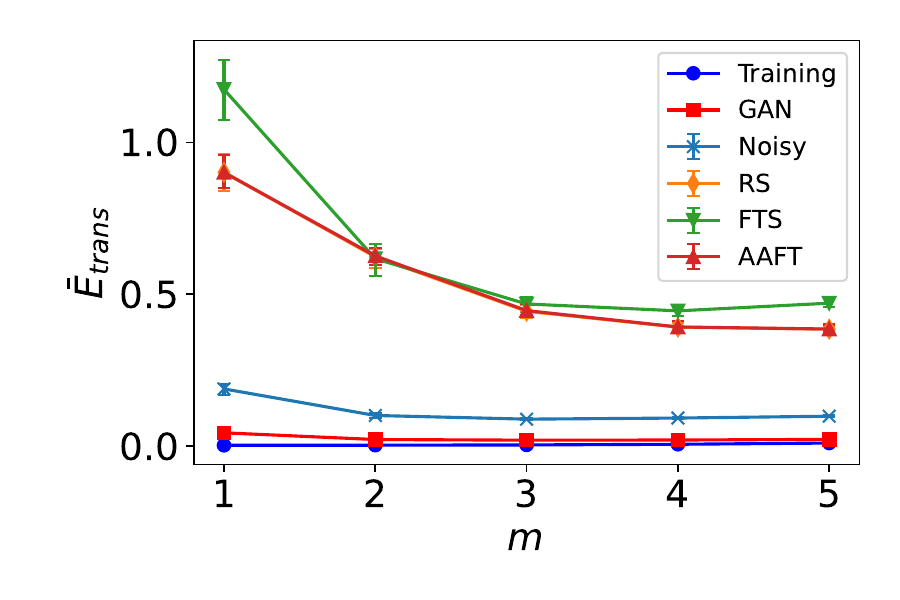}
  \caption{ Comparison of translation errors between the original (training), generated, 
  noisy, and surrogate time series. The abscissa is the embedding dimension. AAFT, amplitude-adjusted Fourier transform; FT, Fourier transform; RS, random shuffle.}
  \label{fig-wayland}
\end{figure}

For the training data and GAN-generated data, we computed $\bar{E}_{\text{trans}}$ using $100,000$-length time series.
For the noisy data and the three types of surrogate data, we created 100 same-length time series based on the training data, and calculated $\bar{E}_{\text{trans}}$ for each sample. Figure~\ref{fig-wayland} shows values of $\bar{E}_{\text{trans}}$ under different $m$ and conditions.
For the noisy and surrogate data, error bars indicate the maximum and minimum values over $100$ trials.
The $\bar{E}_{\text{trans}}$ for the generated data was much lower than that for the surrogate data and notably lower than the noisy data, indicating that the determinism of GAN-generated time series is close to that of the training data.

\subsection{Estimation of maximal Lyapunov exponents}

To test whether the GAN-generated data exhibit sensitive dependence on initial conditions, a defining characteristic of chaotic dynamics, we applied the maximal Lyapunov exponent estimation algorithm~\cite{kantz1994} to the generated time series and compared the result with the training, noisy, and surrogate data, as in the previous section.

In this algorithm, for a reference point $\mathbf{x}_{t_{0}}$ in the embedding space, let $\mathcal{U}_{\varepsilon}(\mathbf{x}_{t_{0}})$ be a set of neighbor points in a sphere of radius $\varepsilon$ centered at the reference point.
Then, the logarithm of average distances between the trajectory from the reference point and the trajectories from the neighborhood points are calculated as a function of relative time step $k$. We repeated this calculation with all points as reference points, taking the average as
% REVISE fix equation
\begin{multline*}
  S(k) = \frac{1}{N} \sum_{t_{0}=1}^{N} \ln \Big[ \frac{1}{|\mathcal{U}_{\varepsilon}(\mathbf{x}_{t_{0}})|} \\
   \sum_{\mathbf{x}_{t_j} \in u_{\varepsilon}(\mathbf{x}_{t_{0}})} |x_{t_{0}+m-1+k} - x_{t_{j}+m-1+k} | \Big],
\end{multline*}
% \begin{multline*}
%   S(k) = \frac{1}{N} \sum_{t_{0}=1}^{N} \ln \Big[ \frac{1}{|\mathcal{U}_{\varepsilon}(\mathbf{x}_{t_{0}})|} \\
%    \sum_{\mathbf{x}_{t_j} \in u_{\varepsilon}(\mathbf{x}_{t_{0}})} \|\mathbf{x}_{t_{0}+k} - \mathbf{x}_{t_{j}+k}\| \Big],
% \end{multline*}
where $N=L-m+1-k$ is the number of reference points.
If the trajectory has initial value sensitivities where the small deviation evolves exponentially, this $S(k)$ should grow linearly with $k$ over a certain range. Its slope is then the estimate of the maximal Lyapunov exponent $\lambda$.
We generated 100 time series of length 100,000 for each condition and estimated the maximal Lyapunov exponents.
%REVISE remove this sentence
% We set the embedding dimension to $m=2$, which from the results of the previous section we consider sufficient.
%%%% added these sentences (8/2)
% If the neighborhood radius $\varepsilon$ is too large, the accuracy of divergence will be reduced, while if it is too small, neighborhood points will not be found.
% Therefore, it is necessary to set an appropriate value. We set $\varepsilon$ to $0.001, 0.01, 0.05, 0.1,$ and $0.1$ when $m$ is $1, 2, 3, 4,$, and $5$ respectively.
Appropriately setting neighborhood radius $\varepsilon$ is crucial; a large one decreases estimation accuracy, and a small one fails to find points. 
We set $\varepsilon$ to $0.001,0.01,0.05,0.1,$ and $0.1$ when $m$ is $1,2,3,4,$ and $5$, respectively.

\begin{figure}[t]
  \centering
  \includegraphics[scale=1.08]{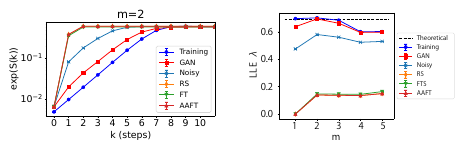}
  \caption{ Estimations of largest Lyapunov exponents. Left:  Slope of $S(k)$ for each condition. 
  %Right: Estimated largest Lyapunov exponents. AAFT, amplitude-adjusted Fourier transform; FT, Fourier transform; RS, random shuffle.
  Right: Estimated largest Lyapunov exponents for each $m$. AAFT, amplitude-adjusted Fourier transform; FT, Fourier transform; RS, random shuffle.
  }
  \label{fig-lle}
\end{figure}

% REVISE Modified paragraphs
% Figure~\ref{fig-lle} (left) shows the variation of $S(k)$ with $k$ for each condition. The error bars indicate the maximum and the minimum of $S(k)$ over $100$ trials. For GAN-generated time series, $S(k)$ increases almost linearly with $k$ in a range of about 1 to 6, similar to the training data, indicating exponential expansion of nearby trajectories, a characteristic of chaotic dynamical systems. The linear increase of $S(k)$ is clearer in the GAN data than in the noisy or surrogate data. In all conditions, $S(k)$ is saturated until $k=7$, where $S(k)$ approaches 1, the system size. Figure~\ref{fig-lle} (right) shows the estimated Lyapunov exponents by estimating the slope of $S(k)$ in the range of $k$ from 1 to 5 using the least-squares method. 
% The figure shows means and standard deviations for 100 trials under each condition. The dotted line is the theoretical value ($\ln{2}$). The GAN data were closer to the training data than the other conditions, and also closer to the theoretical values.
As a representative example, we show the variation of $S(k)$ with $k$ for $m=2$ (Fig.~\ref{fig-lle} left).
%  From the results of the test of determinism in Fig.~\ref{fig-wayland}, the determinism is sufficiently high for $m=2$, which indicates that the embedding dimension is sufficient. 
The slopes for other $m$ are described in \cite{github}. 
%ここからは同じ
Error bars show max and min $S(k)$ across 100 trials. For GAN-generated series, $S(k)$ increases almost linearly with $k$, similar to the training data, indicating exponential expansion of nearby trajectories, a characteristic of chaotic dynamical systems. This trend is clearer in the GAN series than in the noisy or surrogate series. In all conditions, $S(k)$ is saturated until $k=7$, where $S(k)$ approaches the system size.
Figure~\ref{fig-lle} (right) shows the estimated Lyapunov exponents for $m=1, ..., 5$.
The estimation was done by the least-squares method for the slope of $S(k)$ in the range of $k$ from $1$ to $5$.
% The dots and the error bars indicates means and minimum and maximal for 100 trials under each condition, respectively. 
The dashed line indicates the theoretical value ($\ln{2}$). 
The GAN data were closer to the training data than the other conditions and also closer to the theoretical values in the case of $m=1,2,$ and $3$. When $m=4$ and $5$, deviations from the theoretical values are observed in both training data and GAN data. The reason for this was discussed in \cite{github}. 
% can be considered as follows from the observation of the slope shown in  \cite{github} : when $m$ is large, due to taking a larger neighborhood size $\varepsilon$, the number of steps until the error expanded to the system size shortened, leading to a saturation of the slope.

\subsection{Anomalous transitions in generated time series}
\label{sec-errors}

As Fig.~\ref{fig-returnmap} shows, most points on the return map for the generated time series were near the quadratic curve $(x,f(x))$, indicating that most transitions were close to the true transition process, making errors small. Still, there are a few large deviations, suggesting that large errors rarely appeared in generated time series that deviated significantly from the true trajectory. 
To investigate the statistics of these anomalous transitions, we examined the distribution of the absolute error $d_{t}=|f(x_{t})-x_{t+1}|$.

\begin{figure}[t]
  \centering
  \includegraphics[scale=0.46]{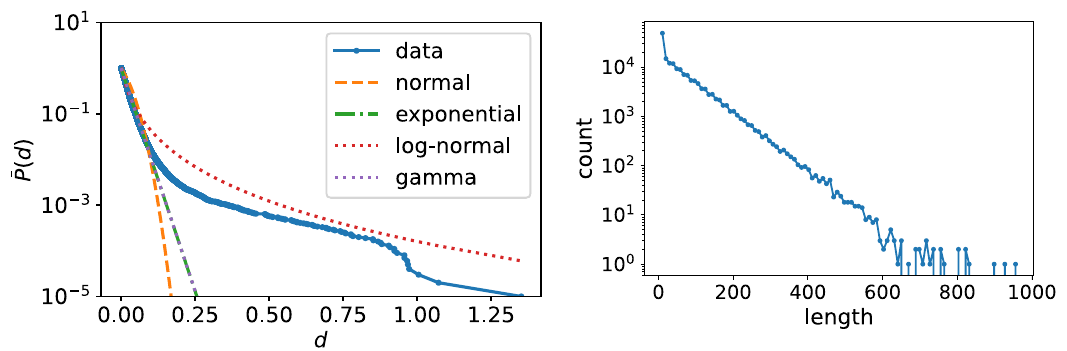}
  \caption{ (Left)~Complementary cumulative distribution of the absolute error $d_{t}$.
            (Right)~Histogram of interval length $l$ between anomalous transitions.}
  \label{fig-errors}
\end{figure}

Using $10^5$ data points, we calculated the complementary cumulative distribution function (CCDF) $\bar{P}(d) = \int_{d}^{\infty}p(d)dd$ of $d$, where  $p(d)$ was the probability density function of $d_{t}$. Figure~\ref{fig-errors} (left) shows the results, using a logarithmic scale on the $y$-axis only.
We also plotted CCDFs for Gaussian, exponential, gamma, and log-normal distributions fitted to the empirical distribution by the maximum likelihood estimation method.
(When fitting the Gaussian distribution, we assumed $d_t$ to be the absolute value of a random variable $e$ that follows a Gaussian distribution with mean 0 and variance $\sigma^{2}_{e}$, where we used the squared mean of $e_t$ as $\sigma^{2}_{e}$. Then, using the probability density function of the Gaussian distribution $p(e;0,\sigma^{2}_{e})$, we calculated the CCDF as $\bar{P}(d) = 1-\int_{-d}^{d}p(e;0,\sigma^{2}_{e})de $.)
The distribution shows an exponential shape at the small-error range. However, there is a deviation from the exponential distribution starting at CCDF values of about $10^{-3}$. This deviation from exponentially decaying distributions indicates that large displacements appear rarely, but much more frequently than expected when following an exponentially decaying distribution. The log-normal distribution also did not fit the empirical distribution. The shape of the empirical distribution is difficult to fit in a single distribution; it is more appropriately described as a mixture or crossover of two distributions.

Next, we examined the statistical distribution of intervals between these anomalous transitions. 
Let $T_{\lambda}=\{t_j | d_{t_j} >\lambda\}$, $(j=1,2,..., n_{\lambda}, t_1<t_2<...,)$ be a set of time steps where $d$ exceeds a threshold $\lambda$, and let  $I = \{l_j =t_{j+1}-t_j\}$ be the set of interval lengths. Figure~\ref{fig-errors}~(right) shows the histogram of $l$ calculated from time series of length $10^5$ with $\lambda=0.1$. Except for very short intervals ($l\leq4$), an exponential distribution well approximates the distribution, implying that the occurrence of anomalous transitions is a Poisson process (i.e., events appear independently of each other). We also confirmed that an exponential distribution can robustly approximate the distribution even when changing the threshold $\lambda$ (not shown).

\section{Discussion} 

We trained a deep convolutional GAN to generate chaotic time series and used nonlinear time series analysis techniques to analyze properties of the generated time series. We found that the determinism of the generated time series was much closer to the original time series than those of random surrogate data or noise-added time series. The maximal Lyapunov exponent estimated from the generated time series was also close to that from the original time series. These results indicate that time series generated by the GAN well retain chaotic characteristics, indicating high capability of the GAN.

However, we observed in the time series some anomalous transitions that deviate significantly from true transitions. Analyses of those errors revealed that their distribution was characterized as a mixture of an exponential distribution representing small errors and another distribution representing large errors that appear infrequently. The distribution of intervals between such anomalous transitions was exponential, which implies that each large error emerged independently.

% REVISE insert (7/31)
Although the GAN's generator is a feed-forward process, when generating the entire time series step by step through a series of operations of accumulated convolution layers, it exchanges local information between nearby temporal points and calculates the feature vector at each time point.
After training, it is expected that the series of feature vectors in the middle layers are gradually shaped so that the neighboring values of the final outputs eventually approximate the mapping relation $(x, f(x), f^{2}(x), ...) $.
From this consideration, we can hypothesize that determinism increases from the initial to final layers when series of feature vectors in middle layers are considered as time series. However, confirming this is for future work.

% Based on this consideration, we can predict that if we treat the series of feature vectors in the intermediate layer as a time series and analyze its determinism in each layer, determinism is expected to be small in the earlier layers, but as we approach the output layer, the determinism increases.   
% The verification of this prediction, however, is beyond the scope of this paper and will be the subject of a future issue.
%%

Our GAN was able to generate largely plausible time series by applying multiple layers of local convolutional operations and nonlinear transformations. However, even such good approximations occasionally break down, producing large errors. Therefore, this study suggests that when using GAN-generated time series for machine-learning tasks like resampling and data augmentation, care should be taken to ensure that anomalous transitions do not adversely affect task performance. A detailed mathematical understanding of these anomalous transitions is a topic for future work.

\acknowledgments

This work was supported by JSPS KAKENHI Grant Numbers JP20K11985 and JP23K11256.

\references


\begin{thebibliography}{9}

\bibitem{kantz2003}
H.~Kantz and T.~Schreiber,
Nonlinear Time Series Analysis,
Cambridge University Press, New York, 2003.

\bibitem{frank2001}
R.~J. Frank, N.~Davey, and S.~P. Hunt,
Time series prediction and neural networks,
J. Intell. Robot. Syst., {\bf31} (2001), 91--103.

\bibitem{jaeger2004}
H.~Jaeger and H.~Haas,
Harnessing nonlinearity: predicting chaotic systems and saving energy in wireless communication,
Science, {\bf 304} (2004), 78--80.

\bibitem{gilpin2021}
W.~Gilpin,
Chaos as an interpretable benchmark for forecasting and data-driven modelling,
in: Proc. NeuIPS Datasets and Benchmarks 2021,
2021.

\bibitem{donahue2018}
C.~Donahue, J.~McAuley, and M.~Puckette,
Adversarial audio synthesis,
in:ICLR 2019, 2019.


\bibitem{zhang2022}
D.~Zhang, M.~Ma, and L.~Xia,
A comprehensive review on GANs for time-series signals,
Neural Comput. Appl., {\bf 34} (2022), 3551--3571.

\bibitem{bai2018}
S.~Bai, J.~Z. Kolter, and V.~Koltun,
An empirical evaluation of generic convolutional and recurrent networks for sequence modeling,
arXiv:1803.01271 [cs.LG].

\bibitem{goodfellow2014} I.~Goodfellow et al.,
Generative adversarial nets,
in: Adv. NIPS, {\bf 27}, (2014)

\bibitem{radford2015}
A.~Radford, L.~Metz, and S.~Chintala,
Unsupervised representation learning with deep convolutional generative adversarial networks,
arXiv:1511.06434 [cs.LG].
% in: ICLR 2016, 2016.

\bibitem{github} \url{https://github.com/yymgch/chaosgan/} (accessed 3, Aug, 2023)

\bibitem{clevert2015}
D.-A. Clevert, T.~Unterthiner, and S.~Hochreiter.
Fast and accurate deep network learning by exponential linear units (ELUs).
arXiv:1511.07289  [cs.LG].


\bibitem{wayland1993}
     R.~Wayland et al.,
     Recognizing determinism in a time series,
 Phys. Rev. Lett., {\bf 70} (1993), 580---582.
     
 \bibitem{theiler1992}
J.~Theiler et al.,
Testing for nonlinearity in time series: the method of surrogate data,
Physica D, {\bf 58} (1992), 77--94.

\bibitem{kantz1994}
H.~Kantz,
A robust method to estimate the maximal lyapunov exponent of a time series,
Phys. Lett. A, {\bf 185} (1994), 77--87.


\end{thebibliography}
\end{document}